\documentclass{article}

\usepackage[preprint,nonatbib]{neurips_2020}

\usepackage[utf8]{inputenc} 
\usepackage[T1]{fontenc}    
\usepackage{hyperref}       
\usepackage{url}            
\usepackage{booktabs}       
\usepackage{amsfonts}       
\usepackage{nicefrac}       
\usepackage{microtype}      
\usepackage{graphicx}
\usepackage{xcolor}
\usepackage{amsmath}
\usepackage{gensymb}
\usepackage{listings}
\usepackage{algorithm}
\usepackage{algorithmic}

\newcommand{\eb}[1]{\scriptsize\,$\pm$\,#1}

\newcommand{\etal}{\textit{et al}.}

\newcommand{\RR}[1]{\raisebox{-0.25mm}{#1}}

\definecolor{codegreen}{rgb}{0,0.6,0}
\definecolor{codegray}{rgb}{0.5,0.5,0.5}
\definecolor{codepurple}{rgb}{0.58,0,0.82}
 
\lstdefinestyle{mystyle}{
    commentstyle=\color{codegreen},
    keywordstyle=\color{magenta},
    numberstyle=\tiny\color{codegray},
    stringstyle=\color{codepurple},
    basicstyle=\ttfamily\footnotesize,
    breakatwhitespace=false,         
    breaklines=true,                 
    captionpos=b,                    
    keepspaces=true,                 
    numbers=left,                    
    numbersep=5pt,                  
    showspaces=false,                
    showstringspaces=false,
    showtabs=false,                  
    tabsize=2
}
 
\title{Milking CowMask for\\ 
Semi-Supervised Image Classification}

\author{
  Geoff French\\
  School of Computing Sciences \\
  University of East Anglia, UK \\
  \texttt{g.french@uea.ac.uk} \\
  \And
  Avital Oliver \\
  Google Research \\
  Brain Tean \\
  \texttt{avitalo@google.com} \\
  \And
  Tim Salimans \\
  Google Research \\
  Brain Tean \\
  \texttt{salimans@google.com}
}

\begin{document}

\maketitle

\begin{abstract}
Consistency regularization is a technique for semi-supervised learning that underlies a number of strong results for classification with few labeled data. It works by encouraging a learned model to be robust to perturbations on unlabeled data. Here, we present a novel mask-based augmentation method called \emph{CowMask}. Using it to provide perturbations for semi-supervised consistency regularization, we achieve a state-of-the-art result on ImageNet with 10\% labeled data, with a top-5 error of 8.76\% and top-1 error of 26.06\%. Moreover, we do so with a method that is much simpler than many alternatives. We further investigate the behavior of CowMask for semi-supervised learning by running many smaller scale experiments on the SVHN, CIFAR-10 and CIFAR-100 data sets, where we achieve results competitive with the state of the art, indicating that CowMask is widely applicable. We open source our code at
\url{https://github.com/google-research/google-research/tree/master/milking_cowmask}.
\end{abstract}

\section{Introduction}
\label{sec:introduction}

Training accurate deep neural network based image classifiers requires large quantities of
training data. While images are often readily available in many problem domains, producing
ground truth annotations is usually a laborious and expensive task that can act as a
bottleneck. Semi-supervised learning offers the tantalising possibility of reducing the
amount of annotated data required by learning from a dataset that is only partially
annotated.

Semi-supervised learning algorithms based on consistency
regularization~\cite{Sajjadi:Mutual, Laine:Temporal, Oliver:RealisticEval} have proved to be simple while effective,
yielding a number of state of the art results over the last few years.
Consistency regularization is driven by encouraging consistent predictions for unsupervised
samples under stochastic augmentation.
Using CutOut \cite{Devries:Cutout} -- in which a rectangular region of an image
is masked to zero -- as the augmentation
has proved to be highly effective, making significant
contributions to the effectiveness of rich augmentation strategies \cite{Xie:UDA, Sohn:FixMatch}.

In this paper, we introduce a simple masking strategy that we call CowMask, whose shapes and appearance
are more varied than the rectangular masks used by CutOut and RandErase \cite{Zhong:RandErase}.
When used to erase parts of an image in a similar fashion to RandErase, CowMask
outperforms rectangular masks in the majority of semi-supervised image classifications tasks that we tested.


We extend the Interpolation Consistency Training (ICT)
algorithm \cite{Verma:ICT} to use mask-based mixing, using both rectangular masks as in CutMix~\cite{Yun:CutMix} and CowMask. Both CutMix and CowMask exhibit strong semi-supervised learning
performance, with CowMask outperforming rectangular mask based mixing in the majority
of cases. CowMask based mixing achieves a state-of-the-art result\footnote{The concurrent SimCLR~\cite{Chen:SimCLR} work achieves a better result using a larger architecture} on semi-supervised Imagenet, and results comparable to state-of-the-art on multiple small image datasets, all without the need for previously proposed complex training procedures.


In Section~\ref{sec:background} we discuss related work that forms the basis of our approach, alongside other semi-supervised learning algorithms for comparison. In Section~\ref{sec:cowmask} we present CowMask, the novel ingredient to our semi-supervised learning algorithm, that is described in Section~\ref{sec:approach}. We present our experiments and results in Section~\ref{sec:experiments}. Finally we discuss our work and conclude in Section~\ref{sec:conclusions}.

\section{Background}
\label{sec:background}

\subsection{Semi-supervised classification}
\label{sec:semisupreview}
A variety of semi-supervised deep neural network image classification approaches have been
proposed over the last several years, including the use of
auto-encoders \cite{Wang:EnAET, Rasmus:LadderNetworks},
GANs \cite{Salimans:ImprovedGANs, Dai:SemiSupBadGAN},
curriculum learning \cite{Cascante:CurriculumSemiSup} and
self-supervised learning \cite{Zhai:S4L}.

Many recent approaches are based on 
consistency regularization~\cite{Oliver:RealisticEval}, a simple approach exemplified by
the $\pi$-model \cite{Laine:Temporal} and the Mean Teacher model \cite{Tarvainen:MeanTeachers}.
Two loss terms are minimized; standard
cross-entropy loss and consistency loss for supervised and unsupervised samples
respectively. Consistency loss measures the difference between predictions resulting
from differently perturbed variants of an unsupervised sample.
The $\pi$-model perturbs samples twice using stochastic augmentation and
minimises the squared difference between class probability predictions. The Mean
Teacher model builds on the $\pi$-model by using two networks; a teacher and a student.
The student is trained using gradient descent as normal while the weights of the teacher
are an exponential moving average of those of the student. The consistency loss term
measures the difference in predictions between the student and the teacher
under different stochastic augmentation.

A variety of types of perturbation have been explored. Sajjadi \etal~\cite{Sajjadi:RegPertSemiSup} employed
richer data augmentation including affine transformations,
while \cite{Laine:Temporal} and \cite{Tarvainen:MeanTeachers} used standard augmentation strategies
such as random crop and noise for small image datasets.
Virtual Adversarial Training (VAT) uses adversarial perturbations that
maximise the consistency loss term.

\subsection{Mixing regularization}

Recent works have demonstrated that blending pairs of images and corresponding ground truths
can act as an effective regularizer. MixUp \cite{Zhang:MixUp} draws a blending
factor from the Beta distribution that is used to interpolate images and ground truth labels.
Interpolation Consistency Training (ICT) \cite{Verma:ICT} extends this approach to work in a
semi-supervised setting by combining it with the Mean Teacher model. The teacher network
is used to predict class probabilities for a pair of images $A$ and $B$ and MixUp is used to
blend the images and the teachers' predictions. The predictions of the student for the blended
image are encouraged to be as close as possible to the blended teacher predictions.

MixMatch \cite{Berthelot:MixMatch} guesses labels for unsupervised samples by sharpening the
averaged predictions from multiple rounds of standard augmentation and blends images and
corresponding labels (ground truth for supervised samples, guesses for unsupervised) using 
MixUp \cite{Zhang:MixUp}. The blended images and corresponding guessed labels are used to
compute consistency loss.

\subsection{Rich augmentation}
AutoAugment \cite{Cubuk:AutoAugment} and RandAugment \cite{Cubuk:RandAugment} are rich augmentation schemes
that combine a number of image operations provided by the Pillow library \cite{Pillow}. AutoAugment learns
an augmentation policy for a specific dataset using re-inforcement learning, requiring a large amount
of computation to do so. RandAugment on the other hand has two hyper-parameters that are chosen via grid
search; the number of operations to apply and a magnitude.

Unsupervised data augmentation (UDA) \cite{Xie:UDA} adds employs a combination of CutOut \cite{Devries:Cutout}
and RandAugment \cite{Cubuk:RandAugment} in a semi-supervised setting achieving state-of-the-art
results in small image benchmarks such as CIFAR-10. Their approach encourages consistency between
the predictions for the original un-modified image and the same image with RandAugment applied.

ReMixMatch \cite{Berthelot:ReMixMatch} builds on MixMatch by adding distribution
alignment and rich data augmentation using CTAugment or RandAugment (depending on the dataset). CTAugment is a variant of AutoAugment
that learns an augmentation policy during training, and RandAugment is a pre-defined set of 15 forms of augmentations with concrete scales. It is worth noting that
ReMixMatch uses predictions from standard 'weak' augmentation as guessed target probabilities
for unsupervised samples and encourages predictions arising from multiple applications
of the richer CTAugment to be close to the guessed target probabilities. The authors found that using
rich augmentation for guessing target probabilities (a la MixMatch) resulted in unstable training.

FixMatch \cite{Sohn:FixMatch} is a simple semi-supervised learning approach that uses standard
'weak' augmentation to predict pseudo-labels for unsupervised samples. The same samples are richly
augmented using CTAugment and cross-entropy loss is computed using the pseudo-labels.
Confidence thresholding \cite{French:SelfEnsDomAdapt} masks the unsupervised
cross-entropy loss to zero for samples whose predicted confidence is below 95\%.

\subsection{Mask-based regularization}

Erasing a rectangular region of an image by replacing it with zeros -- as in Cutout~\cite{Devries:Cutout} --
or noise -- as in RandErase~\cite{Zhong:RandErase} -- has proved to be an effective augmentation strategy
that yields improvements in supervised image classification.

CutOut has proved to be highly effective in semi-supervised classification scenarios.
The UDA authors \cite{Xie:UDA} report impressive results, while the FixMatch authors
\cite{Sohn:FixMatch} report that CutOut alone is as effective as the combination of
the other 14 image operations used in CTAugment.

CutMix \cite{Yun:CutMix} replaces the blending factor in MixUp with a rectangular mask and uses it to mix
pairs of images, effectively cutting and pasting a rectangle from one image onto another. This
yielded significant supervised classification performance gains. CutMix was applied by French \etal as part of a consistency regularization based semi-supervised semantic segmentation algorithm~\cite{French:SemiSupSeg}.

\begin{tabular}{cc}
\begin{minipage}{.8\textwidth}

\begin{algorithm}[H]
\caption{CowMask generation algorithm, with example CowMasks on right with $p=0.5$ and $\sigma \in \{8,16,32\}$.}
\begin{algorithmic}
\REQUIRE mask size $H \times W$
\REQUIRE scale range $(\sigma_{min},\sigma_{max})$
\REQUIRE proportion range $(p_{min},p_{max})$
\REQUIRE inverse error function $\text{erf}^{-1}$
\STATE $\sigma \sim \text{log}\mathcal{U}(\sigma_{min},\sigma_{max})$ \COMMENT{Randomly choose sigma}
\STATE $p \sim \mathcal{U}(p_{min},p_{max})$ \COMMENT{Randomly choose proportion}
\STATE $\mathbf{x} \sim \mathcal{N}^{H \times W}(0, 1)$ \COMMENT{Per-pixel Gaussian noise}
\STATE $\mathbf{x}_s = \text{gaussian\_filter\_2d}(x, \sigma)$ \COMMENT{Filter noise}
\STATE $m = \text{mean}(\mathbf{x}_s)$ \COMMENT{Compute mean and std-dev}
\STATE $s = \text{std\_dev}(\mathbf{x}_s)$
\STATE $\tau = m + \sqrt{2} \cdot \text{erf}^{-1}(2p-1) \cdot s$ \COMMENT{Compute threshold $\tau$}
\STATE $\mathbf{c} = \mathbf{x}_s \leq \tau$ \COMMENT{Threshold filtered noise}
\STATE \textbf{Return} $\mathbf{c}$
\end{algorithmic}
\label{lst:cowmask}
\end{algorithm}
\end{minipage} &

\raisebox{-2mm}{
\begin{minipage}{.13\textwidth}
\fbox{\includegraphics[width=\textwidth]{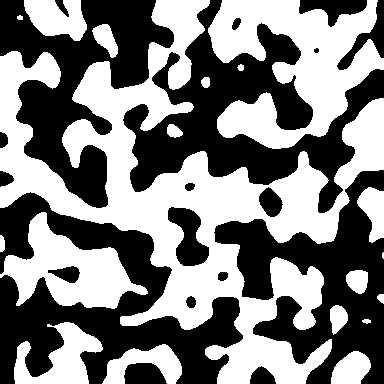}} \linebreak
\fbox{\includegraphics[width=\textwidth]{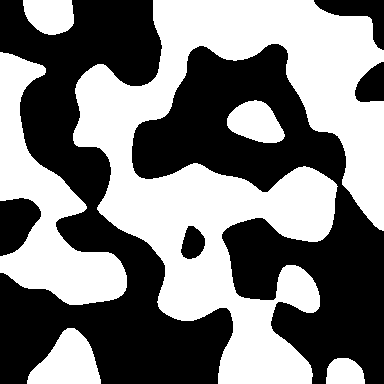}} \linebreak
\fbox{\includegraphics[width=\textwidth]{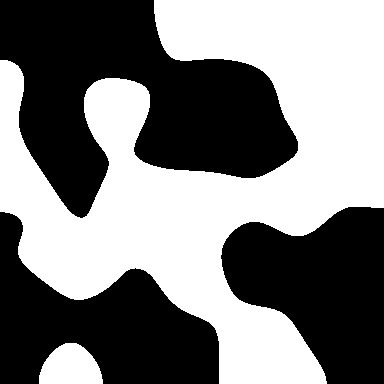}}
\end{minipage}} \\
\end{tabular}

\section{CowMask}
\label{sec:cowmask}

Our exploration of mask-based augmentation for consistency regularization is motivated by the strong performance
of Cutout~\cite{Devries:Cutout} shown in ablation studies in UDA~\cite{Xie:UDA} and FixMatch~\cite{Sohn:FixMatch}.
Furthermore, French \etal~show that semantic segmentation problems exhibit a challenging data distribution where
the cluster assumption -- identified in prior work \cite{Luo:SNTG,Sajjadi:Mutual,Shu:DIRTT,Verma:ICT} as important to the success of consistency regularization -- does not apply. In spite of this, they
obtain strong results using CutMix, suggesting mask-based mixing as a promising avenue.

Here, we propose CowMask; a simple approach to generating the masks shown on the right of Algorithm ~\ref{lst:cowmask}, so called due to its' Friesian cow-like appearance.
We note that the concurrent work FMix \cite{Harris:FMix} uses an inverse Fourier transform to generate masks with a similar visual appearance.

Briefly, a CowMask is generated by applying Gaussian filtering of scale $\sigma$ to normally distributed noise.
A threshold $\tau$ is chosen such that a proportion $p$ of the smooth noise pixels are below $\tau$.
Pixels with a value below $\tau$ are assigned a value of 1, or 0 otherwise.
The scale of the mask features is controlled by $\sigma$ -- as seen in the examples on the right of Algorithm~\ref{lst:cowmask} -- and is drawn from a log-uniform distribution in
the range $(\sigma_{min},\sigma_{max})$. The proportion $p$ of pixels with a value of 1 is drawn from
a uniform distribution in the range $(p_{min},p_{max})$. The procedure for generating a CowMask is provided in Algorithm~\ref{lst:cowmask}.


\section{Semi-Supervised Learning Method}
\label{sec:approach}
We adopt the Mean Teacher\cite{Tarvainen:MeanTeachers} framework as the basis of our approach.
We use two networks; the student $f_\theta(\cdot)$ and the teacher $g_\phi(\cdot)$,
both of which generate class probability vectors.
The student is trained by gradient descent as normal.
After every update to the student, the weights of the teacher are updated to be an exponential moving average of those of the student using $\phi' = \phi\alpha + \theta(1-\alpha)$.
The EMA momentum $\alpha$ controls the trade-off between the stability and the speed at which the teacher follows the student.

Our training set consists of a set of supervised samples $S$ consisting of input images $s$ and corresponding target labels $t$, and a set of unsupervised samples $U$ consisting only of input images $u$.
Given a labelled dataset we select the supervised subset randomly such that it maintains the class balance of the overall dataset\footnote{We use \texttt{StratifiedShuffleSplit} from Scikit-Learn \cite{sklearn_api}} as is standard practice in the literature.
All available samples are used as unsupervised samples. Our models $f_{\theta}$ are then trained to minimize a combined loss:
\[
L = L_\mathcal{S}(f_\theta(s),t) + \omega L_\mathcal{U}(f_\theta(u), g_\phi(u))
\]
where we use standard cross entropy loss for the supervised loss $L_\mathcal{S}(\cdot)$ and consistency loss for the unsupervised loss $L_\mathcal{U}(\cdot)$ that is modulated by the unsupervised loss weight $\omega$.

We explore two different types of mask-based consistency regularization: \emph{mask-based erasure} and \emph{mask-based mixing}. In mask-based erasure we perturb our input data by erasing the part of the input image corresponding to a randomly sampled mask. In mask-based mixing we blend two input images together, with the blending weights given by the sampled mask.
We follow the nomenclature of Cutout and CutMix, using the terms CowOut and CowMix to refer to CowMask
based erasure and mixing respectively.

\begin{tabular}{cc}
\raisebox{0.96cm}{
\begin{minipage}{.45\textwidth}
\begin{algorithm}[H]
\caption{CowOut erasure-based unsupervised loss}
\begin{algorithmic}
\REQUIRE unlabeled image $\mathbf{x}$, CowMask $\mathbf{m}$
\REQUIRE teacher model $g_\phi$
\REQUIRE student model $f_\theta$
\REQUIRE confidence threshold $\psi$
\STATE $\mathbf{\hat{x}} = \text{std\_aug}(\mathbf{x})$ \COMMENT{standard augmentation}
\STATE $\mathbf{z} = \text{stop\_gradient}(g_\phi(\mathbf{\hat{x}}))$ \COMMENT{teacher pred.}
\STATE $q = \max_{i}\mathbf{z}[i] \geq \psi$   \COMMENT{confidence mask}
\STATE $\mathbf{\epsilon} \sim N(0,I)$   \COMMENT{generate noise image}
\STATE $\mathbf{\hat{x}}_{m} = \mathbf{\hat{x}} * \mathbf{m} + \mathbf{\epsilon} * (1 - \mathbf{m})$ \COMMENT{apply mask}
\STATE $\mathbf{y}_{m} = f_\theta(\mathbf{\hat{x}}_{m})$ \COMMENT{student prediction}
\STATE $d = q * ||\mathbf{y}_{m} - \mathbf{z}||^{2}_{2}$   \COMMENT{cons. loss}
\STATE \textbf{Return} $d$
\end{algorithmic}
\label{lst:app:erase_alg}
\end{algorithm}
\end{minipage}} &

\raisebox{-2mm}{
\begin{minipage}{.48\textwidth}
\begin{algorithm}[H]
\caption{CowMix mixing-based unsupervised loss}
\begin{algorithmic}
\REQUIRE unlabeled images $\mathbf{x}_{a}$, $\mathbf{x}_{b}$
\REQUIRE CowMask $\mathbf{m}$
\REQUIRE teacher model $g_\phi$
\REQUIRE student model $f_\theta$
\REQUIRE confidence threshold $\psi$
\STATE $\mathbf{\hat{x}}_{a} = \text{std\_aug}(\mathbf{x}_{a})$ \COMMENT{standard augmentation}
\STATE $\mathbf{\hat{x}}_{b} = \text{std\_aug}(\mathbf{x}_{b})$
\STATE $\mathbf{z}_{a} = \text{stop\_gradient}(g_\phi(\mathbf{\hat{x}}_{a}))$ \COMMENT{teacher pred.}
\STATE $\mathbf{z}_{b} = \text{stop\_gradient}(g_\phi(\mathbf{\hat{x}}_{b}))$
\STATE $c_{a} = \max_{i}\mathbf{z}_{a}[i]$   \COMMENT{confidence of prediction}
\STATE $c_{b} = \max_{i}\mathbf{z}_{b}[i]$
\STATE $\mathbf{\hat{x}}_{m} = \mathbf{\hat{x}}_{a} * \mathbf{m} + \mathbf{\hat{x}}_{b} * (1 - \mathbf{m})$ \COMMENT{mix images}
\STATE $p = \text{mean}(\mathbf{m})$ \COMMENT{scalar mean of mask}
\STATE $\mathbf{z}_{m} = \mathbf{z}_{a} * p + \mathbf{z}_{b} * (1-p)$ \COMMENT{mix tea. preds.}
\STATE $c_{m} = c_{a} * p + c_{b} * (1-p)$ \COMMENT{mix confidences}
\STATE $q = \text{mean}(c_{m} \geq \psi)$   \COMMENT{mean of conf. mask}
\STATE $\mathbf{y}_{m} = f_\theta(\mathbf{\hat{x}}_{m})$ \COMMENT{stu. pred. on mixed image}
\STATE $d = q||\mathbf{y}_{m} - \mathbf{z}_{m}||^{2}_{2}$   \COMMENT{cons. loss}
\STATE \textbf{Return} $d$
\end{algorithmic}
\label{lst:app:mix_alg}
\end{algorithm}

\end{minipage}} \\
\end{tabular}

\subsection{Mask-based augmentation by erasure}
Mask-based erasure can function as an augmentation that can be added to the standard augmentation
scheme used for the dataset at hand, with one caveat.
Similar to prior work \cite{Xie:UDA, Berthelot:ReMixMatch, Sohn:FixMatch} we found it necessary to split
our augmentation into a 'weak' standard augmentation scheme (e.g. crop and flip) and a 'strong' rich
scheme; RandAugment in the case of the prior works mentioned or CowOut in our work.
Weakly augmented samples are passed to the teacher network, generating predictions that are used as pseudo-targets
that the student is encouraged to match for strongly augmented variants of the same samples.
Using 'strong' erasure augmentation to generate pseudo-targets resulted in unstable training.


The $\pi$-model \cite{Laine:Temporal} and the Mean Teacher model \cite{Tarvainen:MeanTeachers} both use a Gaussian ramp-up function to modulate the effect of consistency loss during the early stages of training. Reinforcing the random predictions of an untrained network was found to harm performance. In place of a ramp-up we opt to use confidence thresholding \cite{French:SelfEnsDomAdapt}.
Consistency loss is masked to zero for samples for which the teacher networks' predictions are below a specified threshold. FixMatch \cite{Sohn:FixMatch} uses confidence thresholding for similar reasons.

Our procedure for computing unsupervised consistency loss based on erasure is provided in Algorithm~\ref{lst:app:erase_alg}. For our small image experiments we found that the best value for the unsupervised weight factor $\omega$ is 1.


\subsection{Mask-based mixing}
Alternatively, we can construct an unsupervised consistency loss by mask-based \emph{mixing} of images in place of erasure.
Our approach for mixing image pairs using masks is essentially that of interpolation consistency training (ICT) \cite{Verma:ICT}.
ICT works by passing the original image pair to the teacher network, the blended image to the student, and encouraging the student networks' prediction to match the blended teacher predictions.
Where ICT draws per-pair blending factors a beta distribution, we mix images using a mask, and mix probability predictions with the mean of that mask (the proportion of pixels with a value of 1).

Confidence thresholding required adaptation for use with mix-based regularization.
Rather than applying confidence thresholding to the blended teacher probability
predictions we opted to blend the confidence values before thresholding as this gave
slightly better results. Further improvements resulted from modulating the consistency
loss by the proportion of samples in the batch whose predictions cross the confidence threshold,
rather masking the loss for each sample individually.


The procedure for computing unsupervised mix consistency loss is provided in Algorithm~\ref{lst:app:mix_alg}.
We found that a higher weight $\omega$ was appropriate for mix consistency loss; we used
a value of 30 for our small image experiments.

\section{Experiments and results}
\label{sec:experiments}
We first evaluate CowMix for semi-supervised consistency regularization on the challenging ImageNet dataset, where we match the state of the art. Next, we examine CowOut and CowMix further and compare with previously proposed methods by trying multiple versions of our approach combined with multiple models on three small image datasets: CIFAR-10, CIFAR-100 and SVHN. The training regimes used for both ImageNet and the small image datasets are sufficiently similar that we used the same codebase for all of our experiments.

Our results are obtained by using the teacher network for evaluation. We report our results as error rates presented
as the mean $\pm$ 1 standard deviation computed from the results of 5 runs, each of which uses a different subset of samples as the supervised set. Supervised sets are consistent for all experiments for a given dataset and number of supervised samples.


\subsection{ImageNet 2012}

We contrast the following scenarios: a supervised baseline using 10\% of the dataset, semi-supervised training with the same 10\% of labelled examples using CowMix consistency regularization on all unlabeled examples, and fully supervised training with all 100\% labels.

\subsubsection{Setup}

We used the ResNet-152 architecture. We adopted a training regime as similar as possible to a standard ImageNet ResNet training protocol.
We used a batch size of 1024 and SGD with Nesterov Momentum \cite{Sutskever:Momentum} set to 0.9 and weight decay (via L2 regularization) set to 0.00025. Our standard augmentation scheme consists of inception crop, random horizontal flip and colour jitter, as in~\cite{Tarvainen:MeanTeachers}.We found that the standard learning rate of 0.1 resulted in unstable training, but were able to stabilise it by reducing the learning rate to 0.04~\cite{Tarvainen:MeanTeachers}. We found that our approach benefits from training for longer than in supervised settings, so we doubled the number of training epochs to 180 and stretched the learning rate schedule by a factor of 2, reducing the learning rate at epochs 60, 120 and 160 and reduced it by a factor of 0.2 rather than 0.1. We used a teacher EMA momentum $\alpha$ of 0.999.

We obtained our CowMix results using a mix loss weight of 100 and and a confidence threshold of 0.5.
We drew the CowMask $\sigma$ scale parameter from the range $(32, 128)$.

\subsubsection{Results}

Our ImageNet results are presented in Table~\ref{tab:results:imagenet}.
We match the S$^4$L MOAM \cite{Zhai:S4L} top-5 error result and beat their top-1 error result, with a
simple end-to-end approach and a significantly smaller model.
By comparison the S$^4$L MOAM result is obtained using a 3-stage training and fine-tuning procedure.
The recent SimCLR~\cite{Chen:SimCLR} approach (concurrent work) uses self-supervised contrastive learning followed
by a fine tuning stage.
They beat our result when using a much larger model.
We tested our approach with wider models (e.g. ResNet-50$\times2$) but obtained better results from the deeper
and commonly used ResNet-152.

\begin{table}[h]
\centering
\begin{tabular}{lllll}

\hline
Approach                                        & Architecture          & Params.   & Top-5 err.        & Top-1 err.       \\
\hline
\hline
& \multicolumn{3}{l}{\footnotesize{\textbf{Our baselines}}} \\
Sup 10\%                                        & ResNet-152            & 60M       & 22.12\%           & 42.91\%           \\
Sup 100\%                                       & ResNet-152            & 60M       & 5.67\%            & 21.33\%           \\

\hline
& \multicolumn{3}{l}{\footnotesize{\textbf{Other work}}} \\
Mean Teacher~\cite{Tarvainen:MeanTeachers}      & ResNeXt-152           & 62M       & 9.11\%\eb{0.12}   & --                \\
UDA~\cite{Xie:UDA}                              & ResNet-50             & 24M       & 11.2\%            & 31.22\%           \\
FixMatch~\cite{Sohn:FixMatch}                   & ResNet-50             & 24M       & 10.87\eb{0.28\%}  & 28.54\eb{0.52\%}  \\
S$^4$L Full (MOAM)~\cite{Zhai:S4L}              & ResNet-50$\times$4    & 375M      & 8.77\%            & 26.79\%           \\
SimCLR~\cite{Chen:SimCLR}                       & ResNet-50             & 24M       & 12.2\%            & --                \\
SimCLR                                          & ResNet-50$\times$2    & 94M       & 8.8\%             & --                \\
SimCLR                                          & ResNet-50$\times$4    & 375M      & \bf7.4\%          & --                \\
\hline
& \multicolumn{3}{l}{\footnotesize{\textbf{Our results}}} \\

CowMix                                          & ResNet-152            & 60M       & 8.76\eb{0.07\%}   & \bf26.06\eb{0.17\%}  \\

\hline
\end{tabular}
\caption{Results on ImageNet with 10\% labels. Note that $S^4L$ involves three steps with different training procedures, while CowMix involves a single training run. SimCLR is able to beat CowMix, but only when using a very large model.}
\label{tab:results:imagenet}
\vspace{-1.5em}
\end{table}

\subsection{Small image experiments}
Alongside CowOut and CowMix we implemented and evaluated Mean Teacher, CutOut/RandErase and CutMix, and we compare our method against these using the CIFAR-10, CIFAR-100, and SVHN datasets.

We note the following differences between our implementation and those of CutOut and CutMix: 1. Our boxes are chosen so that they entirely fit within the bounds of the mask, whereas CutOut and CutMix use a fixed or random size respectively and centre the box anywhere within the mask, with some of the box potentially being outside the bounds of the mask. 2. CutOut uses a fixed size box, CutMix randomly chooses an area but constrains the aspect ratio to be that of the mask, we choose both randomly.


\subsubsection{Setup}

For the small image experiments we use a 27M parameter Wide ResNet 28-96x2d with shake-shake regularization \cite{Gastaldi:ShakeShake}. We note that as a result of a mistake in our implementation we used a $3\times3$ convolution rather than a $1\times1$ in the residual shortcut connections that either down-sample or change filter counts, resulting in a slightly higher parameter count.

The standard Wide ResNet training regime \cite{Zagoruyko:WRN} is very similar to that used for ImageNet.
We used the optimizer, but with weight decay of 0.0005 and a batch size of 256.
As before, the standard learning rate of 0.1 had to be reduced to ensure stability, this time to 0.05.
The small image experiments also benefit from training for longer; 300 epochs instead of the standard 200 used in supervised settings. The adaptations made to the Wide ResNet learning rate schedule were nearly identical to those
made to the ImageNet schedule. We doubled its length and reduced the learning rate by a factor
of 0.2 rather than 0.1. We did however remove the last step; the learning rate is reduced at epochs 120 and
240 rather than epochs 60, 120 and 160 as used in supervised settings.
For erasure experiments we used a teacher EMA momentum $\alpha$ of 0.99 and for mixing experiments we used 0.97.

When using CowOut and CowMix we obtained the best results when the CowMask scale parameter $\sigma$ is drawn from
the range $(4, 16)$. We note that this corresponds to a range of $(\tfrac{1}{8}, \tfrac{1}{2})$ relative to the $32\times32$ image size and that the $\sigma$ range used in our ImageNet experiments bears a nearly identical relationship to the $224\times224$ image size used there. For erasure experiments using CowOut we obtained the best results when drawing $p$; the proportion of pixels that are retained from the range $(0.25, 1)$. Intuitively it makes sense to retain at least 25\% of the image pixels as encouraging the network to predict the same result for an image and a blank space is unlikely to be useful. For mixing experiments using CowMix we obtained the best results when drawing $p$ from the range $(0.2, 0.8)$.

We performed hyper-parameter tuning on the CIFAR-10 dataset using 1,000 supervised samples and evaluating on 5,000 training samples held out as a validation set. The best hyper-parameters found were used as-is for CIFAR-100 and SVHN.

\begin{table*}[t]
\begin{center}%
\renewcommand{\arraystretch}{1.150000}%
\resizebox{\linewidth}{!}{%
\setlength{\tabcolsep}{3mm}%
\begin{tabular}{@{ }llllllllll@{ }}
\hline
\RR{Labeled samples}
    & \RR{\bf 40}
    & \RR{\bf 50}
    & \RR{\bf 100}
    & \RR{\bf 250}
    & \RR{\bf 500}
    & \RR{\bf 1000}
    & \RR{\bf 2000}
    & \RR{\bf 4000}
    & \RR{\bf ALL} \\
\hline
\hline

\multicolumn{10}{c}{\footnotesize{\textbf{Other work: uses smaller Wide ResNet 28-2 model with 1.5M parameters}}}\\

EnAET          &              & \bf16.45\%       & \bf\hphantom{0}9.35\% & \hphantom{0}7.6\%\eb{0.34}  & \hphantom{0}7.27\%               & \hphantom{0}6.95\%               & \hphantom{0}6.0\%                &\hphantom{0}5.35\%              &  \\ 
UDA            &              &       &      & \hphantom{0}8.76\%\eb{0.90}      & \hphantom{0}6.68\%\eb{0.24}      & \hphantom{0}5.87\%\eb{0.13}      & \hphantom{0}5.51\%\eb{0.21}      & \hphantom{0}5.29\%\eb{0.25}      &  \\ 
MixMatch       &              &       &      & 11.08\%\eb{0.87}     & \hphantom{0}9.65\%\eb{0.97}      & \hphantom{0}7.75\%\eb{0.32}      & \hphantom{0}7.03\%\eb{0.15}      & \hphantom{0}6.24\%\eb{0.06}      &  \\ 
ReMixMatch     & 14.98\%\eb{3.38} &       &      & \hphantom{0}6.27\%\eb{0.34}      &                      & \hphantom{0}5.73\%\eb{0.16}      &                      & \hphantom{0}5.14\%\eb{0.04}      &  \\ 
FixMatch (RA)  & \bf13.81\%\eb{3.37} &       &      & \bf\hphantom{0}5.07\%\eb{0.65}      &                      &                      &                      & \hphantom{0}4.26\%\eb{0.05}      &  \\ 

\hline

\multicolumn{10}{c}{\footnotesize{\textbf{Other work: uses 26M parameter models}}}\\

EnAET          &              &       &      &                      &                      &                      &                      & \hphantom{0}4.18\%\eb{0.04}      &  \bf\hphantom{0}1.99\% \\ 
UDA            &              &       &      &                      &                      &                      &                      & \hphantom{0}3.7\% / \bf2.7\%      &  \\ 
MixMatch       &              &       &      &                      &                      &                      &                      & \hphantom{0}4.95\%\eb{0.08}      &  \\ 

\multicolumn{10}{c}{\footnotesize{\textbf{Our results: uses 27M parameter Wide ResNet 28-96x2d with shake-shake}}}\\
Supervised            &              & 76.01\%\eb{1.53}      & 69.74\%\eb{2.09}      & 58.41\%\eb{1.60}      & 47.12\%\eb{1.78}      & 36.61\%\eb{1.11}      & 24.53\%\eb{0.80}      & 14.81\%\eb{0.43}      & \hphantom{0}3.57\%\eb{0.09} \\ 

&\multicolumn{9}{l}{\footnotesize{Augmentation / erasure based regularization}}\\
Mean teacher          &              & 75.68\%\eb{3.72}      & 67.77\%\eb{4.17}      & 47.95\%\eb{4.52}      & 29.72\%\eb{5.74}      & 14.14\%\eb{0.56}      & \hphantom{0}8.79\%\eb{0.16}      & \hphantom{0}6.92\%\eb{0.15}      & \hphantom{0}3.04\%\eb{0.07} \\ 
RandErase             &              & 74.67\%\eb{2.13}      & 62.86\%\eb{3.61}      & 37.63\%\eb{7.20}      & 19.22\%\eb{3.34}      & 11.87\%\eb{0.73}      & \hphantom{0}7.05\%\eb{0.14}      & \hphantom{0}5.27\%\eb{0.17}      & \hphantom{0}2.59\%\eb{0.10} \\ 
CowOut                &              & 72.55\%\eb{3.80}      & 56.72\%\eb{3.90}      & 28.45\%\eb{7.03}      & 14.00\%\eb{1.84}      & \hphantom{0}8.98\%\eb{1.11}      & \hphantom{0}6.27\%\eb{0.40}      & \hphantom{0}4.97\%\eb{0.12}      & \hphantom{0}2.50\%\eb{0.10} \\ 

&\multicolumn{9}{l}{\footnotesize{Mix based regularization}}\\
ICT                   &              & 80.08\%\eb{2.57}      & 72.96\%\eb{4.46}      & 44.92\%\eb{7.85}      & 17.10\%\eb{2.15}      & 10.40\%\eb{0.63}      & \hphantom{0}7.75\%\eb{1.23}      & \hphantom{0}5.97\%\eb{0.11}      & \hphantom{0}3.45\%\eb{0.06} \\ 
CutMix                &              & 66.06\%\eb{15.82}      & 34.05\%\eb{6.19}      & \hphantom{0}9.01\%\eb{3.60}      & \hphantom{0}6.81\%\eb{1.04}      & \hphantom{0}5.44\%\eb{0.39}      & \hphantom{0}4.62\%\eb{0.15}      & \hphantom{0}4.11\%\eb{0.19}      & \hphantom{0}2.78\%\eb{0.14} \\ 
CowMix                &              & \bf55.46\%\eb{15.23}      & 23.00\%\eb{3.95}      & \hphantom{0}7.56\%\eb{0.94}      & \hphantom{0}\bf5.34\%\eb{0.80}      & \hphantom{0}\bf4.73\%\eb{0.37}      & \hphantom{0}\bf4.13\%\eb{0.16}      & \hphantom{0}3.61\%\eb{0.07}      & \hphantom{0}2.56\%\eb{0.06} \\ 
\hline

\end{tabular}%
}%
\vspace*{-1mm}%
\caption{Results on CIFAR-10 test set, error rates as $mean \pm std-dev$ of 5 independent runs.}
\label{tab:results:cifar10}
\end{center}
\end{table*}

\begin{table*}[t]
\begin{center}%
\renewcommand{\arraystretch}{1.150000}%
\resizebox{\linewidth}{!}{%
\setlength{\tabcolsep}{3mm}%
\begin{tabular}{@{ }lllllllll@{ }}
\hline
\RR{Labeled samples}
    & \RR{\bf 40}
    & \RR{\bf 100}
    & \RR{\bf 250}
    & \RR{\bf 500}
    & \RR{\bf 1000}
    & \RR{\bf 2000}
    & \RR{\bf 4000}
    & \RR{\bf ALL} \\
\hline
\hline

\multicolumn{9}{c}{\footnotesize{\textbf{Other work: uses smaller Wide ResNet 28-2 model with 1.5M parameters}}}\\

EnAET          &              & 16.92\%      & \hphantom{0}\bf3.21\%\eb{0.21}  & \hphantom{0}\bf3.05\% & \hphantom{0}2.92\%         &  \hphantom{0}\bf2.84\%  & \hphantom{0}2.69\%                    &   \\ 
UDA            &              &      &                      &                      & \hphantom{0}2.55\%\eb{0.99}                     &                      &   &  \\ 
MixMatch       &              &              & \hphantom{0}3.78\%\eb{0.26}     & \hphantom{0}3.64\%\eb{0.46} & \hphantom{0}3.27\%\eb{0.31}     &  \hphantom{0}3.04\%\eb{0.13}  & \hphantom{0}2.89\%\eb{0.06}                    & \\ 
ReMixMatch     & \bf3.55\%\eb{3.87} & \hphantom{0}3.10\%\eb{0.50}      &                      & \hphantom{0}2.83\%\eb{0.30}      &                      & \hphantom{0}\bf2.42\%\eb{0.09}      &  \\ 
FixMatch (RA)  & 3.96\%\eb{2.17}     &      & \bf\hphantom{0}2.48\%\eb{0.38}      &                      & \hphantom{0}\bf2.28\%\eb{0.11}      & & &  \\ 

\hline

\multicolumn{9}{c}{\footnotesize{\textbf{Other work: uses 26M parameter models}}}\\
EnAET          &              &      &                      &                      & \hphantom{0}2.42\%                     &                      &   &  \\ 

\multicolumn{9}{c}{\footnotesize{\textbf{Our results: uses 27M parameter Wide ResNet 28-96x2d with shake-shake}}}\\
Supervised     &              & 71.24\%\eb{5.40}      & 37.02\%\eb{6.15}      & 18.85\%\eb{1.49}      & 11.71\%\eb{0.55}      & \hphantom{0}8.23\%\eb{0.38}      & \hphantom{0}6.01\%\eb{0.46}      & \hphantom{0}2.82\%\eb{0.08} \\ 

&\multicolumn{8}{l}{\footnotesize{Augmentation / erasure based regularization}}\\
Mean teacher   &              & 62.16\%\eb{10.92}     & \hphantom{0}8.23\%\eb{4.62}      & \hphantom{0}3.84\%\eb{0.15}      & \hphantom{0}3.75\%\eb{0.10}      & \hphantom{0}3.61\%\eb{0.15}      & \hphantom{0}3.47\%\eb{0.12}      & \hphantom{0}2.73\%\eb{0.04} \\ 
RandErase      &              & 52.55\%\eb{22.03}     & \hphantom{0}7.61\%\eb{1.71}      & \hphantom{0}6.17\%\eb{1.25}      & \hphantom{0}4.81\%\eb{0.46}      & \hphantom{0}3.66\%\eb{0.15}      & \hphantom{0}3.21\%\eb{0.22}      & \hphantom{0}2.36\%\eb{0.04} \\ 
CowOut         &              & 66.66\%\eb{19.71}     & 12.11\%\eb{1.82}      & \hphantom{0}5.94\%\eb{0.38}      & \hphantom{0}4.36\%\eb{0.29}      & \hphantom{0}3.59\%\eb{0.25}      & \hphantom{0}3.04\%\eb{0.04}      & \hphantom{0}2.42\%\eb{0.09} \\ 

&\multicolumn{8}{l}{\footnotesize{Mix based regularization}}\\
CutMix         &              & \hphantom{0}\bf9.54\%\eb{2.53}     & \hphantom{0}5.62\%\eb{0.93}      & \hphantom{0}4.32\%\eb{0.52}      & \hphantom{0}3.79\%\eb{0.41}      & \hphantom{0}3.26\%\eb{0.27}      & \hphantom{0}2.92\%\eb{0.09}      & \hphantom{0}2.29\%\eb{0.09} \\ 
CowMix         &              & \hphantom{0}9.73\%\eb{4.01}     & \hphantom{0}3.59\%\eb{0.30}      & \hphantom{0}3.80\%\eb{0.32}      & \hphantom{0}3.72\%\eb{0.60}      & \hphantom{0}3.13\%\eb{0.11}      & \hphantom{0}2.90\%\eb{0.19}      & \hphantom{0}2.18\%\eb{0.06} \\ 
\hline

\end{tabular}%
}%
\vspace*{-1mm}%
\caption{Results on SVHN test set, error rates as $mean \pm stdev$ of 5 independent runs.}
\label{tab:results:svhn}
\end{center}
\end{table*}

\begin{table}[h]
\centering
\setlength{\tabcolsep}{2mm}%
\begin{tabular}{@{ }lllll@{ }}
\hline
\RR{\bf \# Labels}
    & \RR{\bf 1000}
    & \RR{\bf 5000}
    & \RR{\bf 10000}
    & \RR{\bf ALL} \\
\hline
\hline

\multicolumn{5}{c}{\footnotesize{\textbf{Other work: uses 1.5M parameters Wide ResNet 28-2 }}} \\

EnAET                 & 58.73\%               & 31.83\%               & 26.93\%\eb{0.21}      & 20.55\%          \\
MixMatch              &                       &                       & 25.88\%\eb{0.30}      &                  \\ 
FixMatch              &                       &                       & \bf22.60\%\eb{0.12}   &                  \\ 

\hline

\multicolumn{5}{c}{\footnotesize{\textbf{Other work: uses 26M parameter models}}}\\

EnAET                 &                       &                       & 22.92\%               & 16.87\%          \\ 

\multicolumn{5}{c}{\footnotesize{\textbf{Our results: 27M param WRN 28-96x2d}}}\\
Supervised            & 78.80\%\eb{0.22}      & 49.24\%\eb{0.40}      & 36.04\%\eb{0.26}      & 18.82\%\eb{0.22} \\ 

&\multicolumn{4}{l}{\footnotesize{Augmentation / erasure based regularization}}\\
Mean teacher          & 76.97\%\eb{0.99}      & 38.90\%\eb{0.48}      & 30.04\%\eb{0.60}      & 17.81\%\eb{0.17} \\ 
RandErase             & 70.48\%\eb{1.05}      & 35.61\%\eb{0.40}      & 28.21\%\eb{0.16}      & 16.71\%\eb{0.29} \\ 
CowOut                & 68.86\%\eb{0.78}      & 38.82\%\eb{0.44}      & 27.54\%\eb{0.29}      & 16.46\%\eb{0.22} \\ 

&\multicolumn{4}{l}{\footnotesize{Mix based regularization}}\\
CutMix                & 64.11\%\eb{2.63}      & 30.15\%\eb{0.58}      & 24.08\%\eb{0.25}      & 16.54\%\eb{0.18} \\  
CowMix                & \bf57.27\%\eb{1.34}   & \bf29.25\%\eb{0.47}   & 23.61\%\eb{0.30}      & \bf15.73\%\eb{0.15} \\

\hline

\end{tabular}%
\caption{Results on CIFAR-100 test set, error rates as $mean \pm stdev$ of 5 independent runs.}
\label{tab:results:cifar100}
\end{table}

\subsubsection{Results}

Our results for CIFAR-10, CIFAR-100 and SVHN datasets are presented in Tables~\ref{tab:results:cifar10}, \ref{tab:results:cifar100} and \ref{tab:results:svhn} respectively. Considering the techniques we explore we find that mix-based regularization outperforms erasure based regularization, irrespective of the mask generation method used. 

We would like to note that our 27M parameter model is larger than the 1.5M parameter models used for the majority of results in other works, so we cannot make an apples-to-apples comparison in these cases. Our CIFAR-10 results are competitive with recent work, except in small data regimes of less than 500 samples where EnAET \cite{Wang:EnAET} and FixMatch \cite{Sohn:FixMatch} outperform CowMix. Our CIFAR-100 and SVHN results are competitive with recent approaches but are not state of the art. We note that we did not tune our hyper-parameters for these datasets.

\section{Discussion}
\label{sec:discussion}

We explain the effectiveness of CowMix by considering the effects of CowMask and mixing based semi-supervised
learning separately.

DeVries \etal~\cite{Devries:Cutout} established that Cutout -- that uses a box shaped mask similar to RandErase - encourages the network to utilise a wider variety of features in order to overcome the varying combinations of parts of an image being present or masked out. In comparison to a rectangular mask the more flexibly shaped CowMask provides more variety and has less correlation between regions of the mask. This increases in the range of combinations of image regions being left intact or erased enhances its effect.

The MixUp~\cite{Zhang:MixUp} and CutMix~\cite{Yun:CutMix} regularizers demonstrated that encouraging network predictions vary smoothly between two images as they are mixed -- using either interpolation or mask-based mixing -- improved supervised performance, with mask-based mixing offerring the biggest gains. We adapted CutMix -- in a similar fashion to ICT -- for semi-supervised learning and showed that mask based mixing yields significant gains when used as an unsupervised regularizer. CowMix adds the benefits of flexibly shaped masks into the mix.

\section{Conclusions}
\label{sec:conclusions}
We presented and evaluated \emph{CowMask} for use in semi-supervised consistency regularization, achieving a new state of the art on semi-supervised Imagenet, with a much simpler method than in previously proposed approaches, using standard networks and training procedures.  We examined both erasure-based and mixing-based augmentation using CowMask, and find that the mix-based variant -- which we call \emph{CowMix} -- is particularly effective for semi-supervised learning. Further experiments on small image data sets SVHN, CIFAR-10, and CIFAR-100 demonstrate that CowMask is widely applicable.

Research on semi-supervised learning is moving fast, and many new approaches have been proposed over the last year alone that use mask-based perturbation. In future work we would like to further explore the use of CowMask in combination with these other recently proposed methods.

\section*{Broader Impact}

Manual annotation is a laborious and expensive task.
This can act as a bottleneck, slowing or preventing the adoption of machine learning systems.
This is particularly likely to affect organisations such as small businesses and non-commercial entities, for which
the financial cost of annotation could act as a hindrance.
Semi-supervised learing offers the potential of reducing this bottleneck, making machine learning accessible to those
with less access to resources.



\bibliographystyle{plain}
\bibliography{milking_cowmask}

\section*{SUPPLEMENTARY MATERIAL -- Illustration of computation}

Here we provide illustrations that give a simplified overview of our approaches.

\subsection*{Mask based erasure}

\begin{figure*}[h]
\centering

\includegraphics[width=.9\textwidth,trim={7mm 95mm 100mm 10mm},clip]{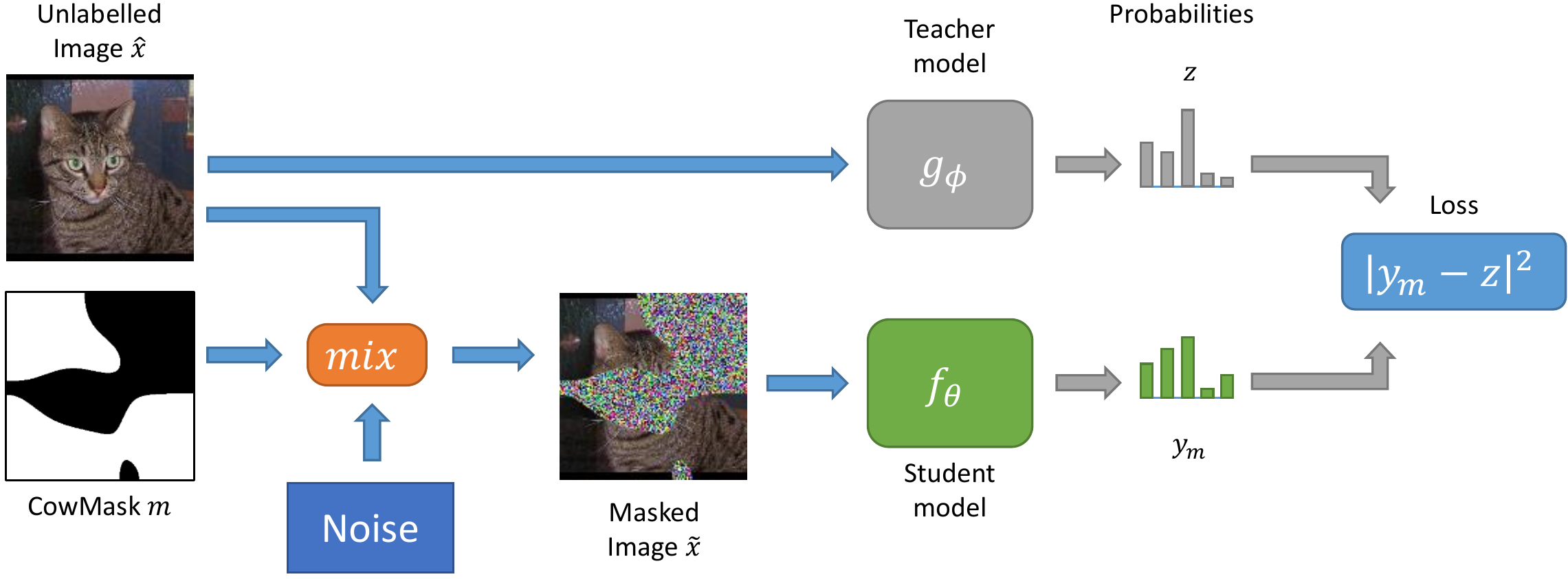}

\caption{\label{fig:semi_sup_erasure}%
Illustration of the unsupervised mask based erasure consistency loss component of semi-supervised image
classification. Blue arrows carry image or mask content and grey arrows carry probability vectors. Note that confidence thresholding is not illustrated here.
}
\vspace{-1em}
\end{figure*}

Our mask-based augmentation by erasure algorithm is illustrated in Figure~\ref{fig:semi_sup_erasure}

 \subsection*{Mask based mixing}

\begin{figure*}[h]
\centering
\includegraphics[width=.9\textwidth,trim={7mm 56mm 55mm 10mm},clip]{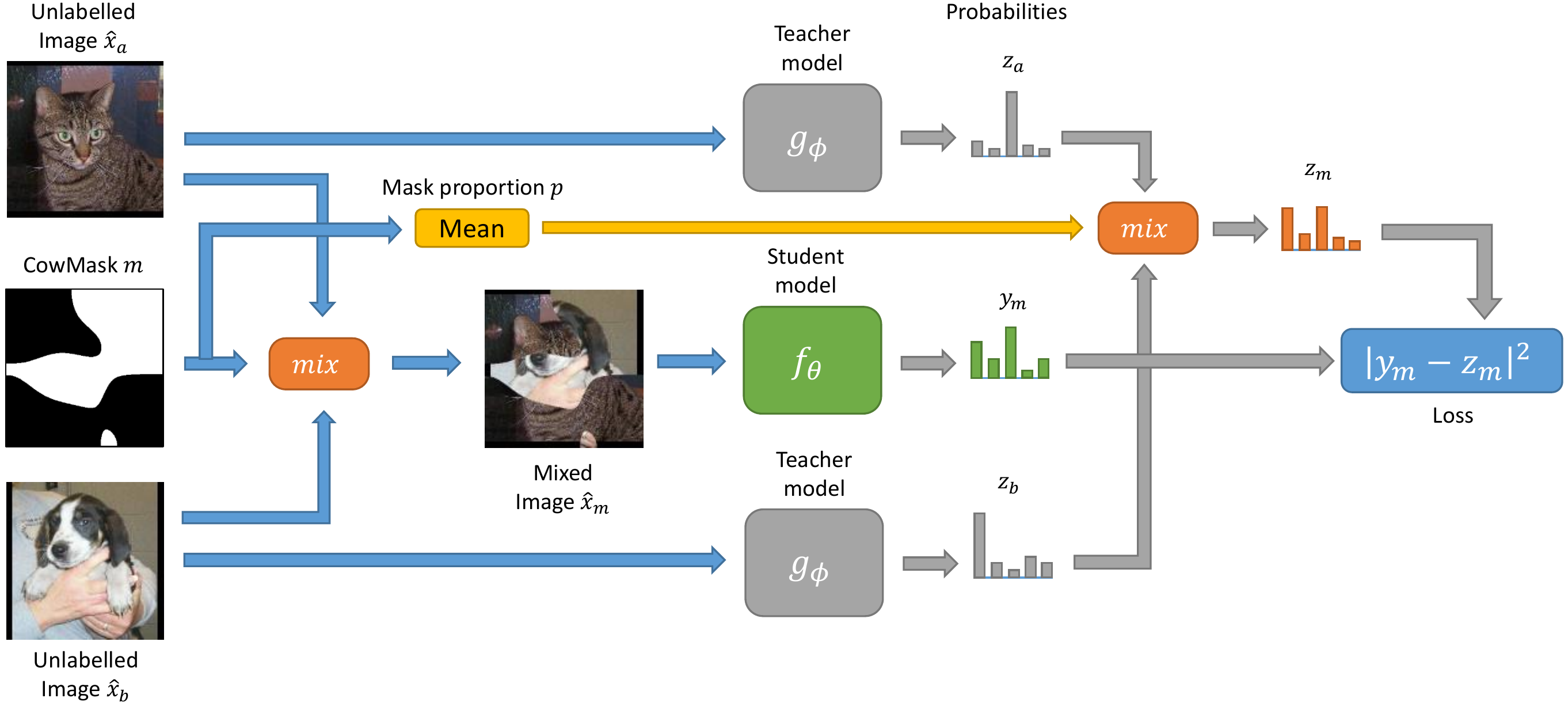}

\caption{\label{fig:semi_sup_mix}%
Illustration of the unsupervised masked based mixing loss component of semi-supervised image classification. Blue arrows carry image or mask content, grey arrows carry probability vectors and yellow carry scalars. Please note that confidence thresholding is not illustrated here.
}
\vspace{-1em}
\end{figure*}

Our mask-based augmentation by mixing algorithm is illustrated in Figure~\ref{fig:semi_sup_mix}.

\end{document}